\documentclass[letterpaper, 10 pt, conference]{ieeeconf}  

\IEEEoverridecommandlockouts                              
\usepackage{graphicx}
\usepackage{balance}
\usepackage{hyperref}       
\usepackage{url}            
\usepackage{booktabs}       
\usepackage{amsfonts}       
\usepackage{nicefrac}       
\usepackage{microtype}      
\usepackage{color}
\usepackage{wrapfig}
\usepackage{graphicx}
\usepackage{epstopdf}
\usepackage{amsmath,amssymb} 
\usepackage{cuted}
\usepackage{bbm}
\usepackage{multirow}
\usepackage{makecell}
\usepackage{multirow}
\usepackage{soul}
\usepackage{arydshln}

\title{\LARGE \bf
Synthetic Data for Multi-Parameter\\ Camera-Based Physiological Sensing 
}

\author{Daniel McDuff$^{1}$, Xin Liu$^{2}$, Javier Hernandez$^{1}$, Erroll Wood$^{3}$ and Tadas Baltrusaitis$^{3}$
\thanks{$^{1}$Daniel McDuff and Javier Hernandez are with Microsoft Research, Redmond, WA, USA
        {\tt\small \{damcduff, javierh\}@microsoft.com}.}%
\thanks{$^{2}$Xin Liu is with the University of Washington, Seattle, WA, USA
        {\tt\small xliu0@cs.washington.edu}.}%
\thanks{$^{3}$Erroll Wood and Tadas Baltrusaitis are with Microsoft, Cambridge, UK
        {\tt\small \{erwood, tabaltru\}@microsoft.com}.}%
}

\begin{document}

\maketitle
\thispagestyle{empty}
\pagestyle{empty}

\begin{abstract}
Synthetic data is a powerful tool in training data hungry deep learning algorithms. However, to date, camera-based physiological sensing has not taken full advantage of these techniques. In this work, we leverage a high-fidelity synthetics pipeline for generating videos of faces with faithful blood flow and breathing patterns. We present systematic experiments showing how physiologically-grounded synthetic data can be used in training camera-based multi-parameter cardiopulmonary sensing. We provide empirical evidence that heart and breathing rate measurement accuracy increases with the number of synthetic avatars in the training set. Furthermore, training with avatars with darker skin types leads to better overall performance than training with avatars with lighter skin types. Finally, we discuss the opportunities that synthetics present in the domain of camera-based physiological sensing and limitations that need to be overcome.
\end{abstract}

  \begin{figure*}[thpb]
      \centering
      \includegraphics[width=\textwidth]{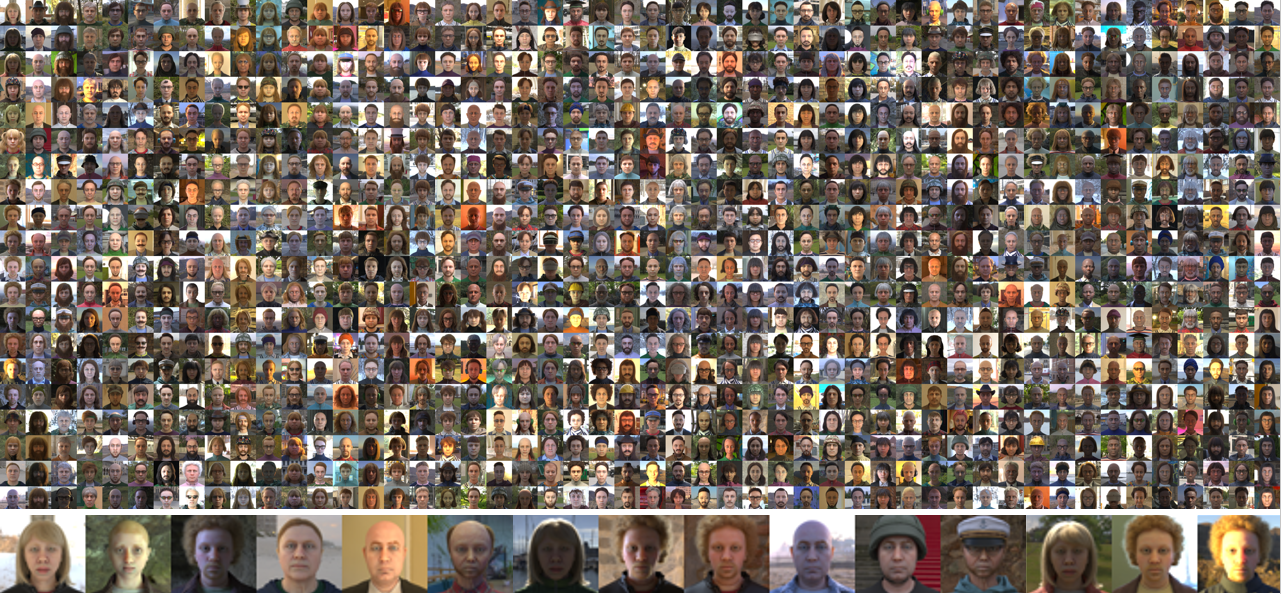}
      \caption{Screenshots of the 1,000 avatars we created for our analyses. Larger examples are shown below to highlight the visual realism of the avatars and the diversity in appearance, lighting and surroundings.}
      \label{fig:faces}
   \end{figure*}

\section{INTRODUCTION}

The application of computer vision in non-contact physiological measurement is an area of growing interest~\cite{mcduff2015survey}. The opportunity to create ubiquitous health sensors out of webcams and smartphones is attractive as it would lower the barrier to measurement and allow for comfortable, convenient longitudinal sensing. Cameras offer the opportunity for capturing rich contextual information, through a myriad of different computer vision models (e.g.,~scene understanding, action and gesture recognition). Context is often tricky to derive from contact-based sensor data (e.g.,~wearables) and is necessary for appropriately interpreting physiological measurements. 

Significant improvements in camera-based measurement accuracy have been achieved using supervised neural models~\cite{chen2018deepphys,liu2020multi}. Training computer vision algorithms to recover physiological signals from video in this way requires synchronized video recordings and gold-standard measurements. Such data is time consuming and expensive to collect. Furthermore, these recordings contain personally identifiable biometric data. In many domains, such as object recognition~\cite{tobin2017domain}, body pose~\cite{shotton2011real} and gaze estimation~\cite{shrivastava2017learning}, face and gesture recognition~\cite{kortylewski2019analyzing} and scene understanding~\cite{handa2016understanding}, synthetic data has become a key tool for training models. Synthetic data pipelines enable systematic data generation with precisely synchronized labels and parameterized control of the dataset properties. A recent paper presented the first example of synthetics applied to the problem of remote imaging photoplethysmography (iPPG) measurement~\cite{mcduff2020advancing}. However, otherwise there remains much to study in this domain.  

Once a synthetics pipeline has been created, it can be used to generate data in a highly scalable fashion, but how much does additional data impact the performance of camera-based physiological sensing systems? Will algorithms benefit from learning from hundreds or thousands of synthetic avatars?
An attractive property of synthetics pipelines is that the data distribution can be controlled. For example, making it easier to sample uniformly from appearance characteristics such as skin type and gender which are known to differently impact the performance of computer vision models~\cite{buolamwini2018gender}. But does controlling the distribution of synthetic data actually lead to an improvement in how the resulting algorithms perform and help create less biased models? And what other insights can we gain from using synthetic data?

In this paper, we present a set of systematic experiments using synthetic data in the training of non-contact vision-based physiological measurement algorithms. First, we examine how the number of synthetically generated avatars in a training set impacts the performance of pulse rate and breathing rate measurement on two benchmark video datasets. Next, we investigate how the skin type distribution of these avatars impacts performance. To achieve this we leverage a state-of-the-art synthetics pipeline for creating video sequences of avatars with physically-grounded simulations of blood flow and breathing. We synthesize 1,000 high-fidelity videos of avatars in this way, to our knowledge the largest such synthetics dataset that exists.

To summarize the contributions of this work are to: 1)~present the first multi-parameter camera-based physiological measurement results using training data created via a synthetics pipeline, 2)~show that generalization of both pulse and breathing measurements from video improves with increased numbers of synthetic avatars in the training set, and 3)~reveal that training on faces with darker skin types appears to improve model generalization across most skin type categories, compared to training on faces with lighter skin types.

\section{BACKGROUND}

\subsection{Vision-Based Physiological Measurement}

Video-based physiological measurement is an established and growing interdisciplinary field of research. Over the past two decades~\cite{wu2000photoplethysmography,blazek2000near,takano2007heart,verkruysse2008remote,poh2010non,de2013robust,wang2015exploiting,chen2018deepphys,liu2020multi,liu2021metaphys} increasingly sophisticated methods have driven significant reductions in error. The field has benefited by grounding these models via a principled approach to modeling the optics of the skin~\cite{Wang2016b}. However, the best performing algorithms are by and large supervised neural networks~\cite{chen2018deepphys,liu2020multi,zhan2020analysis}. These algorithms are ``data hungry'' and can be brittle if only trained on videos that do not reflect the diversity of real-world conditions (appearance, lighting, motion, etc.). Indeed, human appearance and physiology do contain large individual variability making generalization challenging in this domain. Models trained on one dataset and tested on another lead to substantially poorer performance than a model tested on data withheld from the same set as the training data. 
Specific examples of this include over-fitting to the video codec and/or rate factor of the videos in the training set~\cite{nowara2021systematic} and to the skin type of the subjects~\cite{nowara2020meta}.

\subsection{Synthetic Data in Computer Vision}

There is a long history of the use of synthetic data in training and evaluating computer vision systems~\cite{veeravasarapu2015model,veeravasarapu2015simulations,veeravasarapu2016model,vazquez2014virtual,qiu2016unrealcv,mcduff2018identifying,mcduff2019characterizing,haralick1992performance,kortylewski2019analyzing,shotton2011real,handa2016understanding,tobin2017domain,shrivastava2017learning}. 
Synthetics have been employed extensively in models for face and body analysis specifically~\cite{kortylewski2018empirically,kortylewski2019analyzing,mcduff2018identifying,mcduff2019characterizing,baltrusaitis2020high}. But the same is not true for camera-based physiological sensing. 
One attractive feature of synthetic data generation using parameteric models is the ability to control the distribution of samples. These data can then be used to help address problematic biases that exist in models. For example, Kortylewaski et al.~\cite{kortylewski2018empirically,kortylewski2019analyzing} show that the damage of real-world dataset biases on facial recognition systems can be partially addressed by pre-training on synthetic data. 
In this work we built on this prior work and examine specifically what synthetic data can offer camera-based physiological measurement. In particular, we focused on two questions: How much does increasing the number of subjects with different facial appearances improve cross-dataset generalization performance? And how is performance on subjects with different skin types impacted by the distribution of skin types in the synthetic training set. To avoid conflating the impact of real videos with synthetic data we train \emph{only} on synthetics. Previous work shows that we can expect some additional benefit by combining synthetic and real data~\cite{mcduff2020advancing} but that is not our focus here.

\begin{figure*}[t!]
  \includegraphics[width=\textwidth]{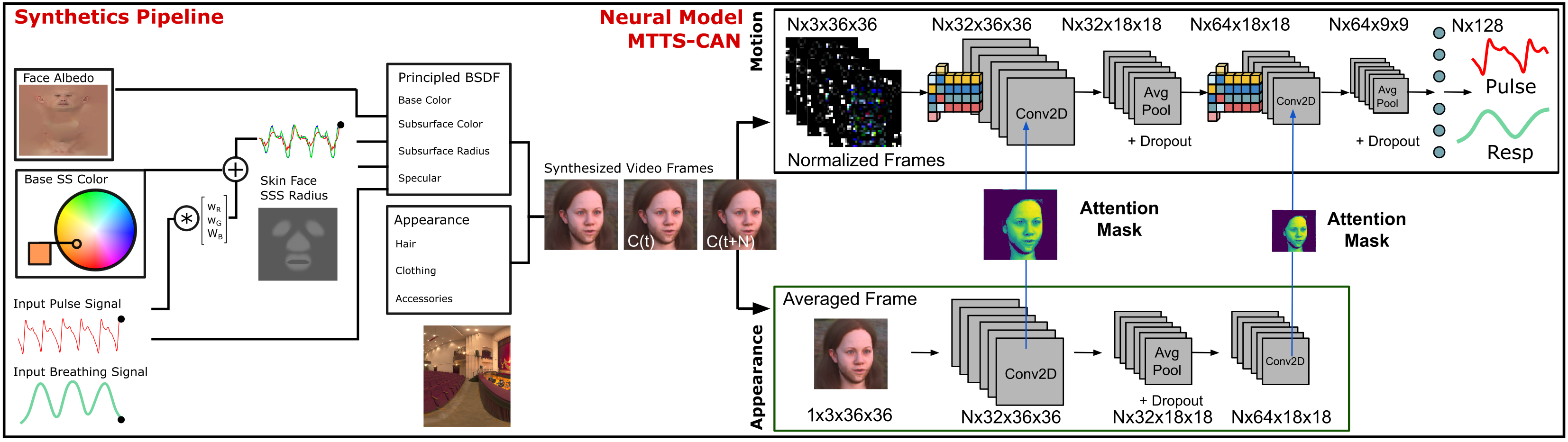}
  \caption{A combined illustration of our multi-parameter data synthesis pipeline and multi-task temporal shift convolutional attention network for camera-based physiological measurement. In this work the models were trained entirely on synthetic data and tested on real videos.}
  \label{fig:mtts-can}
\end{figure*}

\section{SYNTHETIC DATASET}

Generating our synthetic dataset involved creating facial avatars that had simulations of facial blood flow and breathing motions. This section provides more details about the synthetics pipeline. 
We synthesized a large corpora of avatars with unique combinations of facial appearance, head motion and expression, and environment (including ambient lighting configuration). These dimensions were selected based on some of the most significant generalization challenges that existing non-contact physiological models face (e.g.,~generalizing to different appearance, motion and illumination conditions).  

\subsection{Base Avatars}

\textbf{Facial Appearance.} Our first goal was to assess how increasing the number of avatars impacted generalization performance. Therefore, we created 1,000 unique appearances which required randomly picking a skin material with a particular albedo texture. For approximately half of the appearances, we included some form of facial hair (beard and/or moustache).
In addition, we modified the skin color properties of the different faces. Fig.~\ref{fig:faces} shows some examples.

\textbf{Facial Motion}. We introduced rigid head motions by rotating the head about the vertical axis at angular velocities of 0, 10, 20, and 30 degrees/second. In addition, we synthesized videos with smiling, blinking, and mouth opening by using a collection of artist-created blend shapes.

\textbf{Environment.} We randomly picked a static background and illumination on the face~\cite{debevec2006image} from a large existing collection~\cite{zaal2018hdri}. Fig.~\ref{fig:faces} shows some examples.


\subsection{Physiological Changes}

To simulate the facial blood flow, we leveraged photoplethysmographic signals from PhysioNet~\cite{goldberger2000physiobank}. In particular, we used the BIDMC PPG and Respiration Dataset~\cite{pimentel2016toward} which includes 53 8-minute contact PPG and respiration recordings from the original MIMIC-II dataset~\cite{saeed2011multiparameter}. We then added each of the signals to the avatars by modifying two main properties of a physically-based shading material\footnote{https://www.blender.org/}.

\textbf{Subsurface Skin Color:} We simulated skin tone changes by globally varying subsurface color across all skin pixels on the albedo map which is a texture map transferred from a high-quality 3D face scan.

\textbf{Subsurface Scattering:} We manipulated the subsurface radius for the channels to capture the changes in subsurface scattering as the blood volume varies. In particular, we used an artist-created spatially-weighted scattering radius texture (see Fig.~\ref{fig:mtts-can}) which captures variations in the thickness of the skin across the face.

\textbf{Breathing Motion:} We controlled both the torso and head motions to simulate motions of the body due to breathing. Specifically, the pitch of the chest was rotated subtly using the breathing input signal. The rotation of the head was dampened slightly to create greater variance in the appearance changes and so that the breathing motions were most dominant in the chest and shoulders.

\section{TESTING DATASETS}

We performed all our testing on videos of real people with gold-standard measurements for reference. 

\textbf{AFRL}~\cite{estepp2014recovering}: Videos were recorded at 658x492 pixel resolution and 120 frames per second (fps). Twenty-five participants (17 males) were recruited to participate in the study. Gold-standard PPG measurements were captured from a fingertip sensor and breathing measurements were captured from a chest strap. These were recorded using a research-grade biopotential acquisition unit. Each participant was recorded six times for 5-minutes each. The angular velocity of the head motion was increased in each task and this process was repeated twice in front of two background screens.  

\textbf{MMSE-HR}~\cite{zhang2016multimodal}: 
102 videos of 40 participants were recorded at 25 fps capturing 1040x1392 resolution images during spontaneous emotion elicitation experiments. The ground truth contact signal was measured via a Biopac2 MP150 system\footnote{\url{https://www.biopac.com/}} which provided pulse rate at 1000 fps and was updated after each heartbeat. These videos feature smaller but more spontaneous motions than those in the AFRL dataset. Gold-standard breathing measurements are not included in the MMSE-HR dataset.


   \begin{figure}[thpb]
      \centering
      \includegraphics[width=0.7\columnwidth]{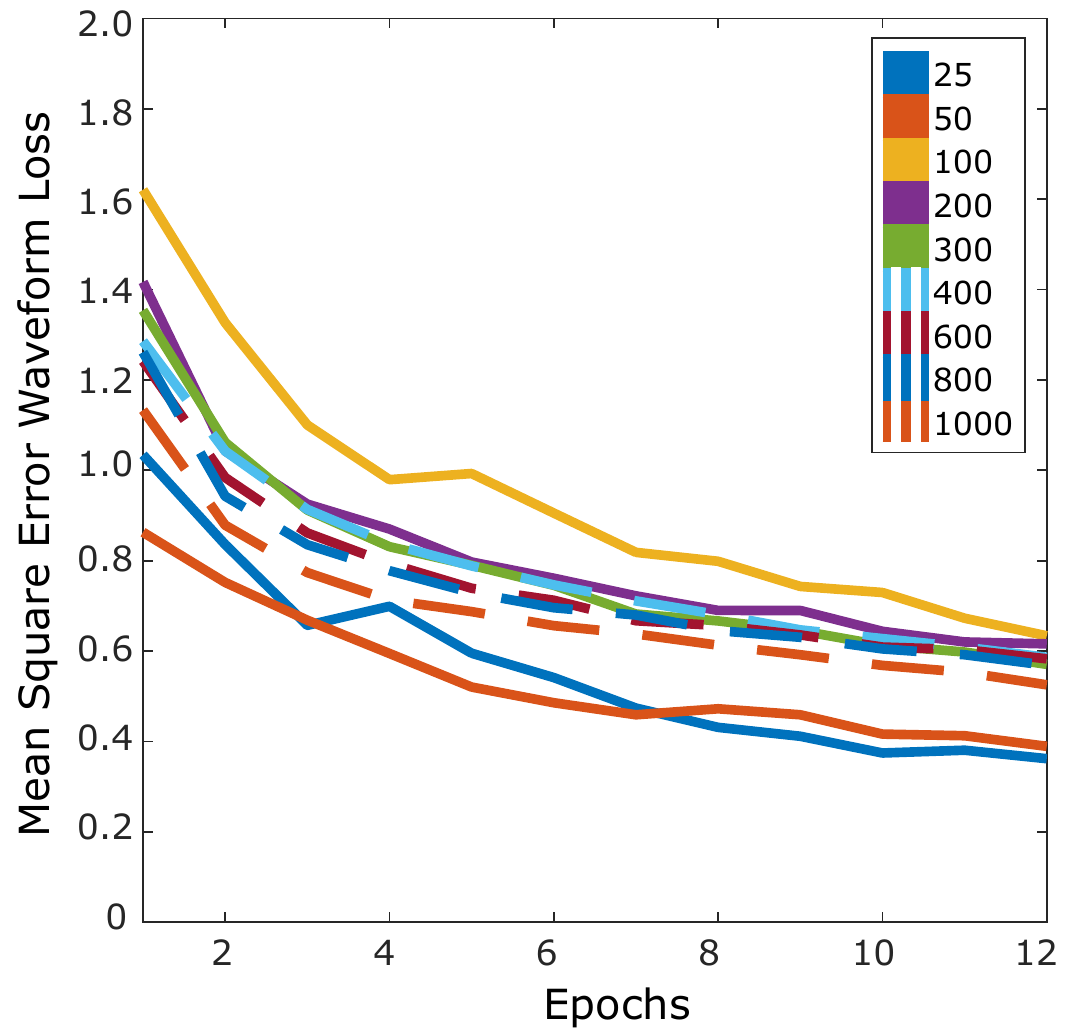}
      \caption{Training loss over 12 epochs for models trained on [25, 50, 100, 200, 300, 400. 600. 800, 1000] avatars.}
      \label{fig:loss}
   \end{figure}

   \begin{figure}[thpb]
      \centering
      \includegraphics[width=89mm]{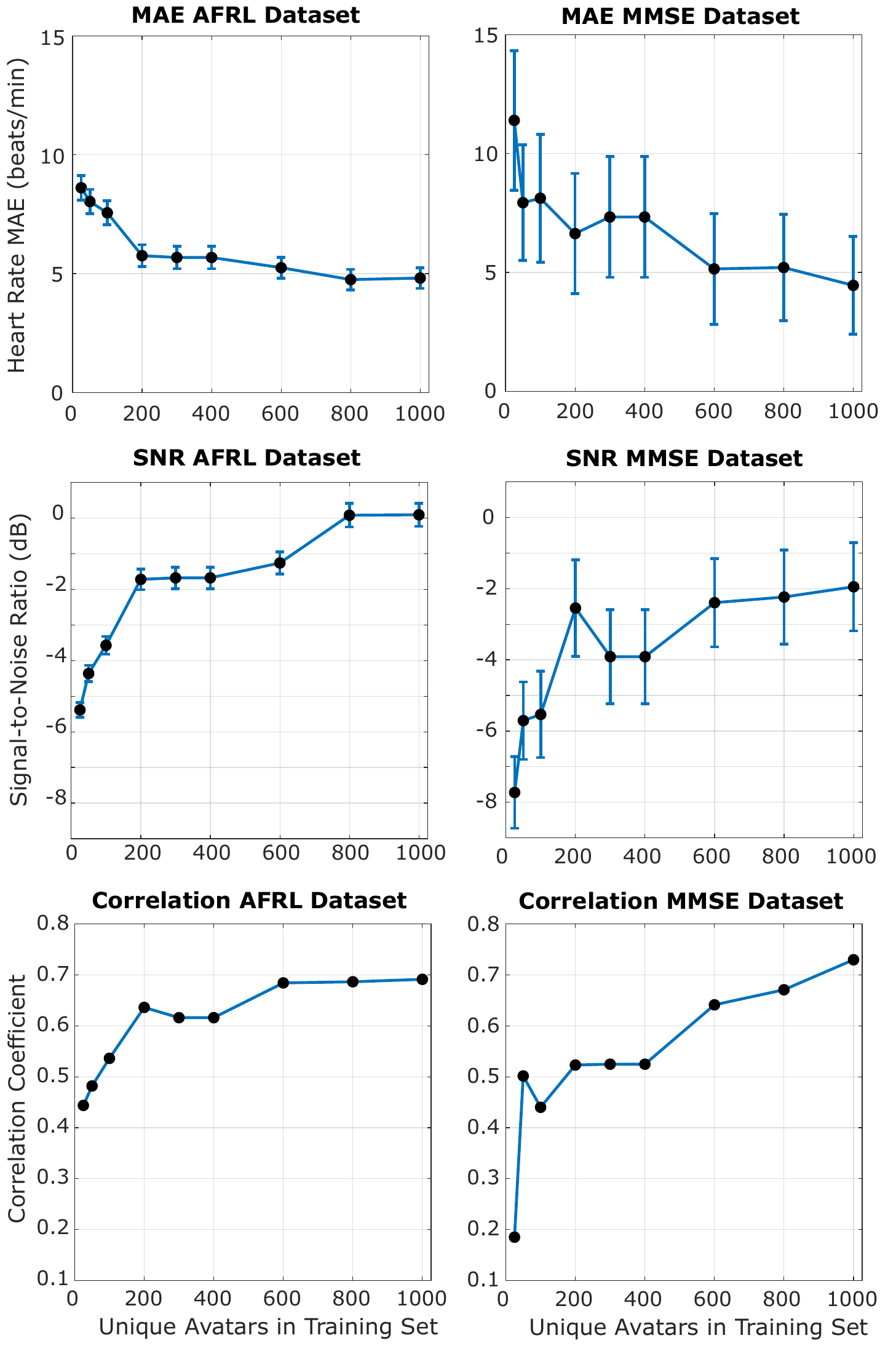}
      \caption{Pulse rate error (beats/min) for models trained with  [25, 50, 100, 200, 300, 400. 600. 800, 1000] randomly sampled avatars. Standard error bars shown for MAE and SNR plots.}
      \label{fig:numberavatars}
   \end{figure}
   
     \begin{figure}[thpb]
      \centering
      \includegraphics[width=47mm]{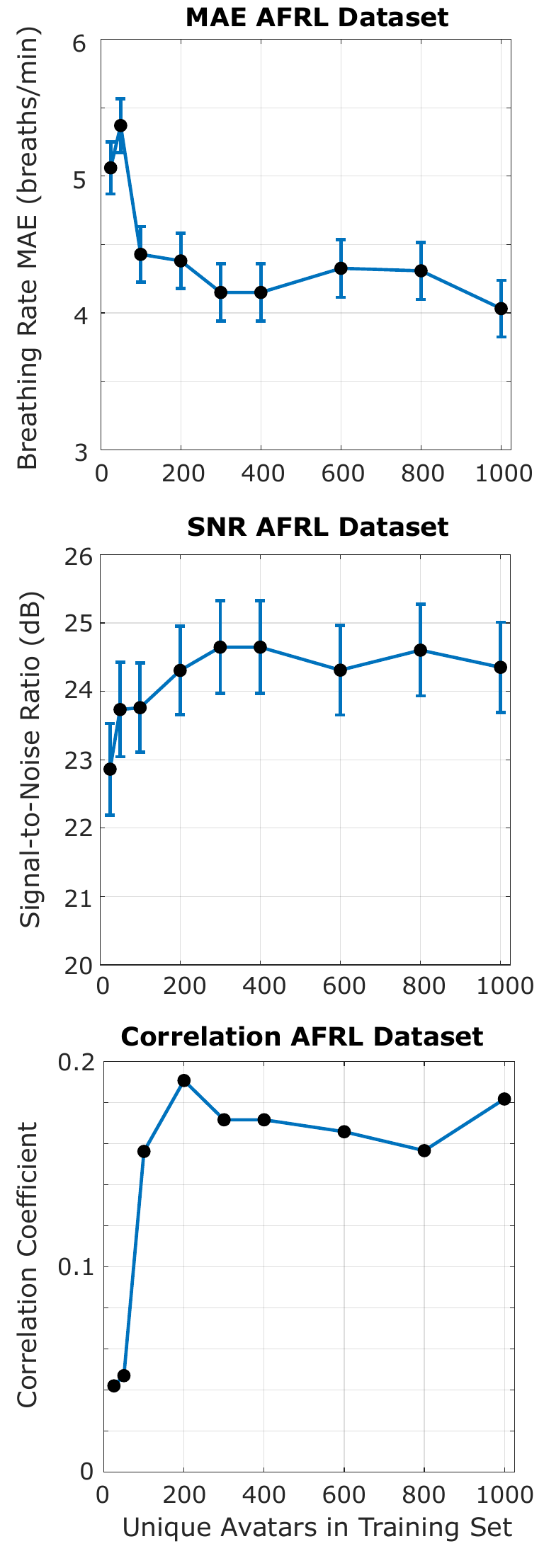}
      \caption{Breathing results for models trained with  [25, 50, 100, 200, 300, 400. 600. 800, 1000] randomly sampled avatars. Standard error bars shown for MAE and SNR plots.}
      \label{fig:numberavatars_resp}
   \end{figure}
   
   \begin{figure}[thpb]
      \centering
      \includegraphics[width=\columnwidth]{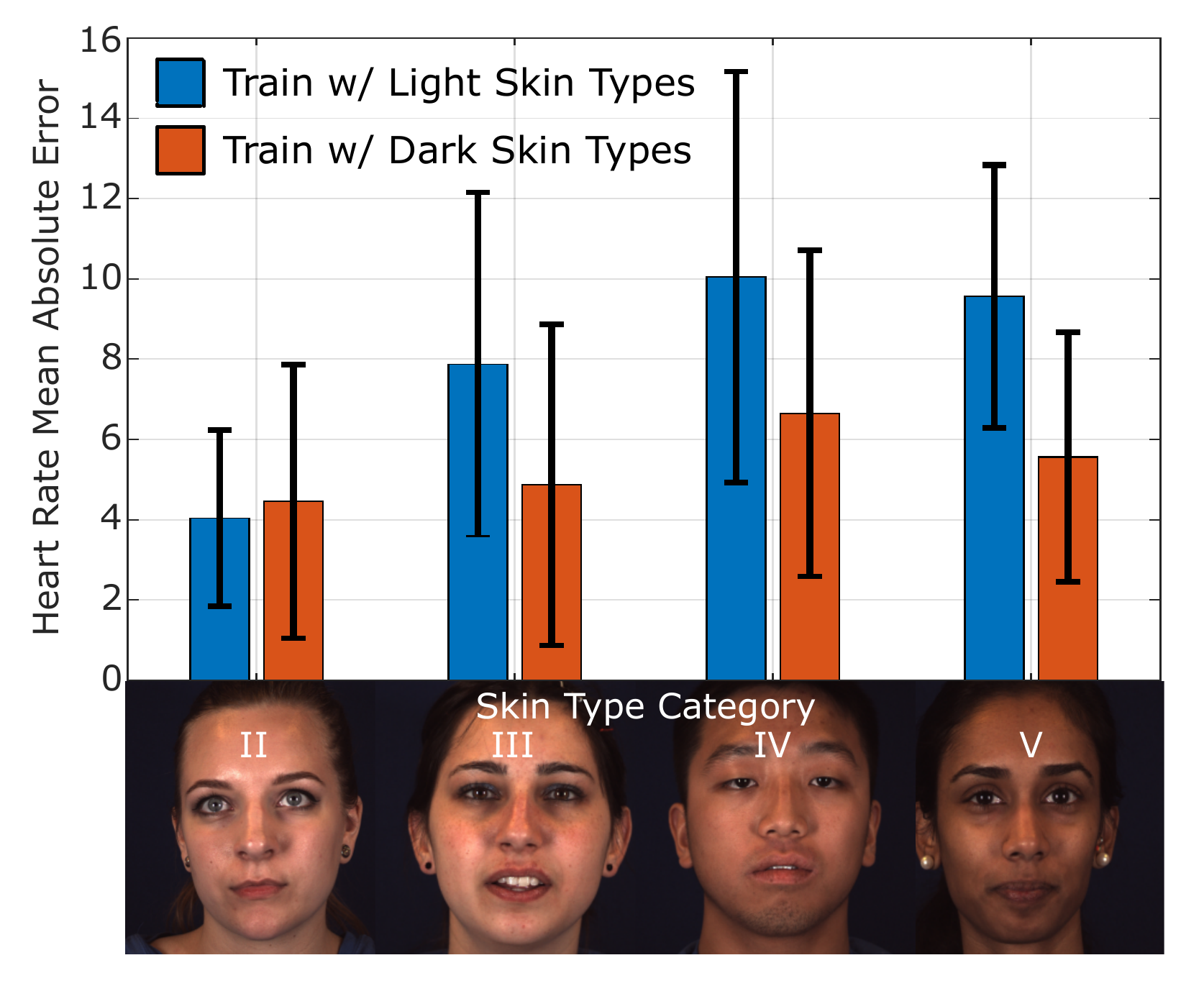}
      \caption{Pulse rate error (beats/min) for models trained with light skin type (blue) and dark skin type (red) avatars.  Results are shown by Fitzpatrick skin type (II, III, IV, V) for subjects in the MMSE-HR dataset. }
      \label{fig:diversityavatars}
   \end{figure}


\section{Model Training and Testing}
Our goal here was not to propose a new inference model, therefore we used an existing state-of-the-art neural architecture for evaluating how synthetic data can improve camera-based physiological sensing. This model, multi-task temporal shift convolutional attention network (MTTS-CAN)~\cite{liu2020multi}, captures rich spatial and temporal relationships in the data and is therefore a good candidate for investigating the impact of the synthetic data. MTTS-CAN has a two-branch network, illustrated in Fig.~\ref{fig:mtts-can}, comprising of a motion branch and an appearance branch. The motion branch efficiently models spatial-temporal features by shifting the frames along the temporal axis. The appearance branch provides an attention mechanism to help guide the motion representation to focus on spatial regions of interests (e.g.,~skin) containing physiological signals instead of others (e.g.,~hair).
The loss function during training was the average mean squared error of pulse and breathing waveform predictions compared to the ground truth waveforms.  We trained using the first-derivative of the waveforms as prior work has shown that this is effective for predicting both PPG and breathing signals from video~\cite{chen2018deepphys}.

Using our synthetics pipeline, we generated videos of 1,000 avatars with blood flow (PPG) and breathing signals. Fig.~\ref{fig:faces} shows a frame from each of the avatar videos. The videos were six seconds in length and were created at a frame-rate of 30Hz, providing 180 frames per avatar and a total of 180,000 frames. This is still a relatively small number of frames compared to many video datasets, but arguably has much greater variability in terms of appearance. Synthesizing face videos is still relatively computationally expensive and time consuming. Creating these high fidelity avatar videos took approximately one month using three computers each with an Nvidia M40 GPU.

For each model in our experiment, we trained the network for a fixed number (12) epochs. Fig.~\ref{fig:loss} shows that although we vary the number of avatars in our experiments, the training loss consistently converges by this point. 
To evaluate the trained models we tested on the AFRL and MMSE datasets. For the AFRL data we divide each video into 5 60-second non-overlapping windows. For the MMSE dataset we use the full video for each prediction. We use three commonly employed evaluation metrics for pulse and breathing: mean absolute error between the predicted rate and the ground truth, signal-to-noise ratio (SNR), and correlation between predicted rate and the ground truth.

\section{SYNTHETICS IN TRAINING}
\subsection{Number of Subjects in the Training Set}

First, we tested models in which we trained networks with data from [25, 50, 100, 200, 300, 400, 600, 800, 1000] avatars. We randomly sample from the pool of avatars to create the subsets for training. Fig.~\ref{fig:numberavatars} shows the mean absolute pulse rate error, SNR and correlation for the AFRL and MMSE datasets (error bars show standard error).  Fig.~\ref{fig:numberavatars_resp} shows the mean absolute breathing rate error, SNR and correlation for the AFRL dataset (the MMSE dataset does not include breathing ground truth data). In both cases (pulse rate and breathing rate estimation) the testing mean absolute error dropped significantly for networks trained with 600 avatars compared to 25 avatars. 
For pulse rate measurement we saw diminishing returns with more than 600 avatars and with breathing rate measurement we saw diminishing returns with more than 200 avatars. These numbers probably reflect the limits of the variance in samples that can be created with out current simulation. After generating 600 unique avatars any additional avatars we generated were likely to have some similarities with one or more of the avatars in the set, thus these additional training samples offered little additional new information. We plan to develop our synthetics pipeline further to address this bottleneck.

\subsection{Sim-to-Real Gap}

From the results in Figs.~\ref{fig:numberavatars} and~\ref{fig:numberavatars_resp} it was apparent that even when training with 1000 avatars the heart rate and breathing rate measurement performance was still not comparable with the state-of-the-art~\cite{liu2020multi,liu2021metaphys,mcduff2020advancing}. We hypothesize that this is due to the ``sim-to-real'' gap.  Training only on synthetic data has limitations because there remains a domain gap between the appearance of synthetic avatars and real videos. Combining our synthetic data with real videos will likely lead to improvements~\cite{mcduff2020advancing} as would other methods for addressing ``sim-to-real'' generalization.  Our goal here was to primarily to examine the impact of synthetic dataset properties and therefore, we leave these steps for future work.

\subsection{Diversity of Subjects in the Training Set}

Our synthetic data allowed us to examine empirically how the distribution of appearance of the avatars impacts generalization performance. Perhaps the most obvious appearance characteristic when measuring the blood volume pulse optically is skin type. Previous work has found systematic biases in the performance of non-contact photoplethymography measurement algorithms with skin type~\cite{addison2018video,nowara2020meta}. Specifically, performance is over signficantly poorer for darker skin types.
The albedo textures were sorted by skin tone. We then trained two models one with ``lighter'' skin tone avatars and another with ``darker'' skin tone avatars.  Fig.~\ref{fig:diversityavatars} shows the pulse rate mean absolute error (beats/min) for the two models on the MMSE-HR dataset by Fitzpatrick skin type~\cite{fitzpatrick1988validity}. We expected to observe that the model trained on light skin type avatars would perform best on light skin type participants (groups II and III) and the model trained on darker skin type avatars would perform best on dark skin type participants (groups IV and V). However, we observed that the model trained on dark skin type avatars performed as well as, or better than the other model on all categories. This suggests that training on data with darker skin types leads to a more robust model, perhaps because the task is harder - forcing the model to learn better representations or a more robust attention mechanism. This is an interesting observation that warrants further investigation.

\section{Discussion}

Our results highlight that synthetic data can be used to train multi-parameter camera-based physiological measurement algorithms. Our experiments have shown that increasing the diversity of appearance of avatars can positively impact generalization performance on real videos for both pulse and breathing measurement. A rich synthetics pipeline presents the possibility to exploit this further, by generating data with more varied facial expressions, body motions, and illumination conditions. In addition to leveraging these at training time, they could also be used for systematic testing. 
However, it is clear that there is a gap between the performance that can be obtained when training only on our current simulation data.

\subsection{Opportunities for Synthetics}

Synthetics open up a number of promising directions for camera-based physiological measurement. As shown in this work, the controlled generation of different facial attributes and potential variations (e.g.,~motion, environment) allowed us to systematically study the effect of different factors and create more generalizable models. In addition, this methodology allowed us to gain more understanding about the impact of each of the variations in a standardized setting which could be used for prioritizing research questions such as ``what is the source of variation that most heavily impacts the algorithm?.'' The generation of synthetic data enabled us to quickly increase the volume of data. This is likely be even more advantageous for training even more data hungry models such as transformers. Finally, the proposed methodology offers the potential to improve other relevant physiological domains. For instance, this work considered data from healthy individuals to generate the physiological signals but future work may consider leveraging datasets with cardio-respiratory abnormalities that may be difficult to replicate during data collection from real people, especially in the context of camera-based physiological sensing.

\subsection{Limitations of Synthetics}
Despite the many benefits, it is important to note that the proposed approach has some limitations too. One of the biggest challenges is that the creation of hyper-realistic avatars like the ones considered in this work requires a large overhead both in terms of time and expense. Once the synthetics pipeline is created, generating the avatars themselves is computationally intensive. However, we expect these costs to be reduced in the future. Another limitation is that using synthetic data can only get us so far in terms of performance. Even though we leveraged state-of-the-art avatar generation, there is still some domain gap between real people and synthesized ones. Future efforts may leverage advances in generative networks like StyleGAN~\cite{Karras2019} to further increase their realism and bridge the ``sim-to-real'' gap.

\section{Conclusion}

We have shown, via empirical evidence that increasing the number of unique avatars in a synthetic dataset can lead to reductions in physiological parameter estimation. We hope that this evidence inspires more research using synthetic data for training camera-based physiological sensing algorithms and that in turn we are able to realize more of the potential for this technology.

\balance{}
















\bibliographystyle{IEEEtran}
\bibliography{references}

\end{document}